\documentclass[10pt,twocolumn,letterpaper]{article}

\usepackage[pagenumbers]{cvpr} 

\definecolor{cvprblue}{rgb}{0.21,0.49,0.74}
\usepackage[pagebackref,breaklinks,colorlinks,allcolors=cvprblue]{hyperref}
\newcommand{\whitefootnote}[1]{%
  \begingroup
  \renewcommand\thefootnote{\textcolor{white}{\arabic{footnote}}}%
  \hypersetup{hidelinks}
  \footnote{\textcolor{black}{#1}}
  \endgroup
}

\def\paperID{15162} 
\def\confName{CVPR}
\def\confYear{2026}

\title{Rotate Your Character: Revisiting Video Diffusion Models for High-Quality 3D Character Generation}

\author{
\textbf{Jin Wang$^{*,1,2}$ \quad Jianxiang Lu$^{*,1}$ \quad Comi Chen$^1$ \quad Guangzheng Xu$^1$ \quad Haoyu Yang$^1$ \quad Peng Chen$^1$}\\
\textbf{Na Zhang$^1$ \quad Yifan Xu$^{1}$ \quad Longhuang Wu$^{1}$ \quad Shuai Shao$^1$\quad Qinglin Lu$^{\dagger,1}$\quad Ping Luo$^{2}$}  \\ 
$^1$Hunyuan, Tencent \quad $^2$The University of Hong Kong
}

\begin{document}
\twocolumn[{
\renewcommand\twocolumn[1][]{#1}
\maketitle
\begin{center}
  \vspace{-11pt}
  \includegraphics[width=1.0\linewidth]{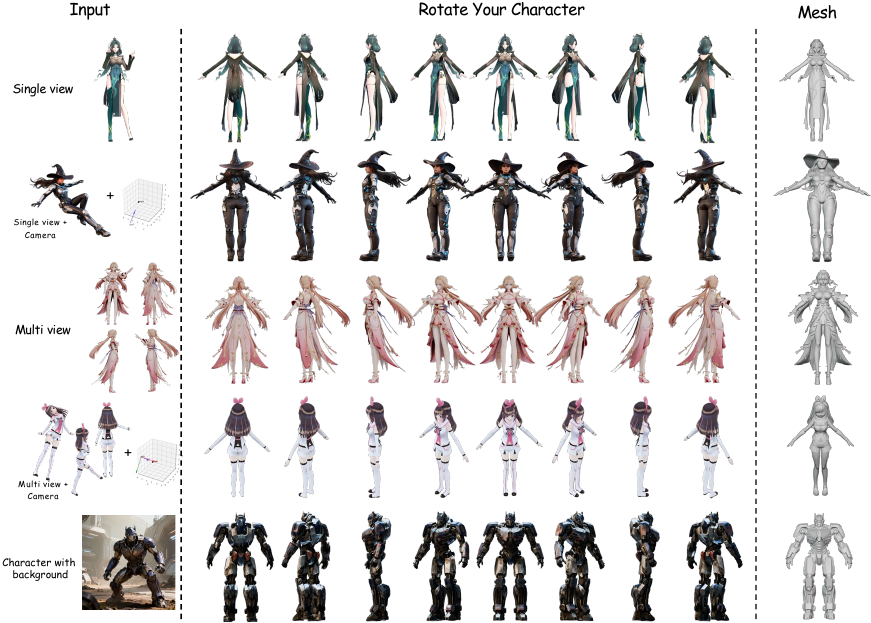}
  \captionsetup{type=figure,font=small}
  \caption{In this paper, we present RCM (\textbf{R}otate your \textbf{C}haracter \textbf{M}odel), an advanced image-to-video diffusion model tailored for high-quality character generations, featuring pose canonicalization given single/multi-view images, high-resolution generations, controllable viewpoints given camera poses and robustness to various background distractions.
  }
  \label{fig:teaser}
\end{center}
}]

\whitefootnote{$^*$ Equal Contribution, $^\dagger$ Corresponding Author.} 

\begin{abstract}
Generating high-quality 3D characters from single images remains a significant challenge in digital content creation, particularly due to complex body poses and self-occlusion. 
In this paper, we present RCM (\textit{\textbf{R}otate your \textbf{C}haracter \textbf{M}odel}), an advanced image-to-video diffusion framework tailored for high-quality novel view synthesis (NVS) and 3D character generation. 
Compared to existing diffusion-based approaches, RCM offers several key advantages: (1) transferring characters with any complex poses into a canonical pose, enabling consistent novel view synthesis across the entire viewing orbit, (2) high-resolution orbital video generation at 1024×1024 resolution, (3) controllable observation positions given different initial camera poses, and (4) multi-view conditioning supporting up to 4 input images, accommodating diverse user scenarios.
Extensive experiments demonstrate that RCM outperforms state-of-the-art methods in both novel view synthesis and 3D generation quality. 
Code and model will be made publicly available.
\end{abstract} 

\section{Introduction}
\label{sec:intro}

From Hollywood blockbusters to immersive VR experiences, converting a single image into a fully realized 3D character has been a long-standing and difficult goal. 
Compared to painstaking manual modeling, such automated solutions significantly slash production time and even open doors for non-experts. Yet, the core challenge persists: inferring comprehensive 3D geometry from flat images demands sophisticated reasoning about the character's hidden sides and spatial structure.

Motivated by the recent success of diffusion models in image synthesis \cite{rombach2022high,esser2024scaling,labs2025flux1kontextflowmatching,flux2024}, researchers have begun exploring how these generative frameworks can be extended to 3D content creation.
Given the scarcity of large-scale 3D character datasets, early approaches sought to distill the knowledge of powerful 2D diffusion models for 3D content generation \cite{poole2022dreamfusion,chen2023fantasia3d,lin2023magic3d,liu2023zero,wang2023prolificdreamer}.
However, such 2D-to-3D lifting methods often suffered from severe multi-view inconsistency—commonly known as the Janus problem—and incurred notably slow generation speeds.
To overcome these limitations, subsequent works extended 2D diffusion models into multi-view generation frameworks \cite{shimvdream,liu2023syncdreamer,shi2023zero123++,liu2023one,liinstant3d,liu2024one,tang2024lgm,peng2024charactergen}, enabling the synthesis of consistent multi-view renderings in a feed-forward manner.
Meanwhile, with the rapid advancement of video diffusion models \cite{lin2024open,zheng2024open,yang2024cogvideox,wan2025wan,kong2024hunyuanvideo}, recent studies have explored leveraging these temporal-consistency frameworks to further enhance cross-view coherence \cite{voleti2024sv3d,zuo2024videomv,yang2024hi3d}.
Nevertheless, under the unique and demanding setting of 3D character generation, these methods continue to face substantial challenges, primarily due to characters’ highly articulated body poses, intricate textures, and the diverse requirements of user-driven scenarios—such as handling multiple conditional input images.

In this paper, we present RCM (\textit{\textbf{R}otate your \textbf{C}haracter \textbf{M}odel}), an advanced image‑to‑video diffusion framework designed for high‑quality novel view synthesis (NVS) and 3D character generation.
Specifically, RCM offers the following key capabilities:
1) \textbf{Pose canonicalization} – given a character image in an arbitrary pose and background, RCM can transfer the character into a standard A/T pose while maintaining consistent appearance throughout a full orbital rotation;
2) \textbf{High‑resolution generation} – RCM produces character videos at a resolution of 1024×1024, ensuring fine detail and visual fidelity;
3) \textbf{Controllable viewpoints} – RCM allows users to specify diverse initial camera poses, providing flexible observation angles of the generated character; and
4) \textbf{Multi‑view conditioning} – RCM supports up to four input images to describe a single character, enhancing consistency and adaptability across varied user scenarios.
These advantages underscore the practicality of RCM in handling complex, industry‑level character generation tasks, as illustrated in Figure \ref{fig:teaser}.

To construct our proposed RCM, we build upon the state‑of‑the‑art video diffusion model Wan 2.2 \cite{wan2025wan} and design a progressive training strategy for effective 3D character generation.
Specifically, the training process is divided into three stages, each dedicated to learning pose canonicalization, viewpoint initialization, and character rotation, respectively.
Moreover, inspired by prior research \cite{he2025cameractrl,li2025hunyuan}, we introduce an external Camera Encoder to capture the subtle variations across different viewpoint conditions.
The Camera Encoder processes Plücker embeddings \cite{sitzmann2021light} derived from camera pose matrices, enabling accurate, geometry‑aware viewpoint conditioning within the diffusion framework.
To further strengthen the generalization capability of RCM, we construct a carefully curated training dataset comprising characters in diverse poses and complex backgrounds. This well‑designed training pipeline enables RCM to robustly handle challenging real‑world cases, producing consistent and high‑quality generations even under significant appearance or pose variations.

In experiments, we conduct comprehensive experiments to assess the effectiveness of our proposed RCM. 
To this end, we establish two challenging benchmarks targeting both in‑the‑wild and hard‑case scenarios. The first benchmark, dubbed RCM-Wild, consists of AI‑generated characters and is evaluated through extensive human perceptual studies, while the second benchmark, dubbed RCM-Hard, includes professionally designed character models and is assessed automatically using objective metrics. Both quantitative results and qualitative comparisons demonstrate the superior performance of our proposed RCM over existing methods, highlighting its strong potential for real‑world 3D character content creation and production pipelines.

Overall, the contributions of this paper are summarized as follows. 
1) We present RCM, an advanced image-to-video diffusion framework that achieves high-quality 3D character generation with pose canonicalization, high-resolution generation, controllable viewpoints, and multi-view conditioning capabilities.
2) We introduce a progressive three-stage training strategy that systematically teaches the model pose normalization, viewpoint initialization, and character rotation, along with a Camera Encoder for precise geometry-aware viewpoint control.
3) We establish two rigorous benchmarks for evaluating character generations, which will be open-sourced to the research community. Through extensive experiments, we show that RCM achieved superior performance on character generations.
\section{Related Work}
\label{sec:relatedwork}
\subsection{Video Diffusion Models}
Recent advances in diffusion models \cite{ho2020denoising,dhariwal2021diffusion,rombach2022high,flux2024} have extended their generative capabilities from static images to dynamic videos, enabling the synthesis of temporally coherent and visually compelling motion sequences. 
Early works \cite{ho2022video,blattmann2023stable,blattmann2023align,chen2024videocrafter2} focused on short video generation by adapting image diffusion backbones with temporal attention modules to model frame‑to‑frame correlations. 
With the emergence of large‑scale video datasets \cite{chen2024panda,wang2024vidprom} and powerful training techniques \cite{lipmanflow,liuflow}, subsequent models such as Sora \cite{openai2024sora}, Open‑Sora \cite{zheng2024open}, CogVideoX \cite{yang2024cogvideox}, Seedance \cite{gao2025seedance}, Wan \cite{wan2025wan}, Veo \cite{google2024veo}, and HunyuanVideo \cite{kong2024hunyuanvideo} have significantly improved video quality, resolution, and duration. 
These state‑of‑the‑art video diffusion models demonstrate strong temporal consistency, better motion understanding, and controllable generation abilities, laying a solid foundation for downstream applications such as editing \cite{wiedemer2025video}, novel view synthesis \cite{voleti2024sv3d} and 3D content generation \cite{yang2024hi3d}.

\subsection{Diffusion Models for 3D Content Generation}
Diffusion-based approaches have recently shown great promise for 3D content creation, leveraging the strong generative priors learned from large‑scale 2D image datasets. Early works \cite{poole2022dreamfusion,chen2023fantasia3d,lin2023magic3d,liu2023zero,wang2023prolificdreamer,zhuhifa,huangdreamtime,tangdreamgaussian,yi2024gaussiandreamer} such as DreamFusion \cite{poole2022dreamfusion}, Fantasia3D \cite{chen2023fantasia3d}, and Magic3D \cite{lin2023magic3d} pioneered the idea of distilling knowledge from 2D diffusion models to optimize 3D representations, typically using score distillation sampling (SDS). 
While these methods marked significant progress, they often suffered from multi‑view inconsistency and slow optimization. 
To address these issues, later methods \cite{shimvdream,liu2023syncdreamer,long2024wonder3d,shi2023zero123++,liu2023one,wang2023imagedream,liinstant3d,liu2024one,tang2024lgm,xu2024instantmesh,peng2024charactergen,xingmirror,lin2024consistent123,hu2024mvd,tang2023MVDiffusion} introduced multi‑view diffusion architectures that generate consistent images across different viewpoints in a feed‑forward manner, such as SyncDreamer \cite{liu2023syncdreamer}, Zero123++ \cite{shi2023zero123++}, and CharacterGen \cite{peng2024charactergen}. 
More recent efforts \cite{voleti2024sv3d,chen2024v3d,melas20243d,zuo2024videomv,yang2024hi3d} have explored integrating temporal consistency from video diffusion models to enhance view coherence and enable smoother viewpoint transition. Despite these advances, generating high‑quality 3D characters remains particularly challenging due to complex articulated poses, intricate surface details, and the need for generalization across diverse user conditions.
\section{Method}
\label{sec:method}

\begin{figure*}[t]
  \centering
   \includegraphics[width=1.0\textwidth]{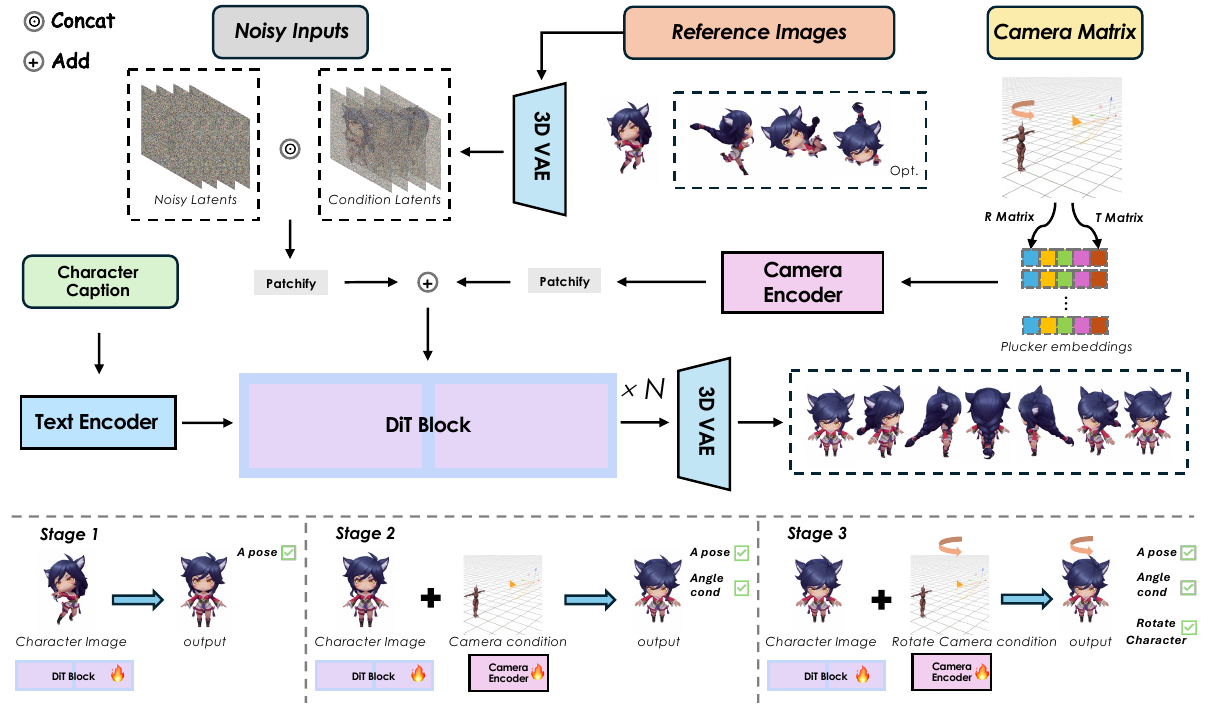}
   \caption{\textbf{Model Architecture and Training Strategies of the Proposed RCM.} The proposed RCM is built upon video diffusion models and incorporates a Camera Encoder to achieve precise viewpoint control. To optimize learning, we adopt a progressive three‑stage training strategy, where each stage focuses respectively on pose canonicalization, viewpoint initialization, and character rotation.}
   \label{fig:mode_arch_train_strategies}
\end{figure*}

\subsection{Preliminaries: Video Diffusion Models}
Video diffusion models extend the classical diffusion framework from static images to the spatio-temporal domain, enabling the generation of coherent video sequences that capture both fine visual detail and smooth motion dynamics. 
Formally, let a video be represented as $\mathbf{x_0} \in \mathbb{R}^{T\times H\times W\times C}$, where $T$ denotes the number of frames, and $(H, W, C)$ denote the height, width, and number of channels. 
Current video diffusion models typically adopt the framework of flow matching~\cite{lipmanflow,liuflow}, defining the forward process as
\begin{equation}
\mathbf{x}_t = (1 - t)\mathbf{x_0} + t\mathbf{x_1}, \quad \mathbf{x_1} \sim \mathcal{N}(0, \mathbf{I}),
\end{equation}
and learning a reverse process parameterized by a neural network $\mathbf{v}_\theta(\mathbf{x}_t, t)$ that predicts the underlying velocity directing the denoising trajectory.

During training, the model receives random samples of $\mathbf{x_0}$ from the data distribution and interpolates them with Gaussian noise $\mathbf{x_1}$ at a randomly sampled timestep $t \in [0,1]$. 
The neural network $\mathbf{v}_\theta$ is then optimized to predict the instantaneous velocity $\mathbf{v} = \mathbf{x_1} - \mathbf{x_0}$ that guides the sample toward the clean data, minimizing the following loss:
\begin{equation}
\mathcal{L}_{\text{FM}}(\theta) = 
\mathbb{E}_{t, \mathbf{x_0}, \mathbf{x_1}}
\left[\|\mathbf{v}_\theta(\mathbf{x}_t, t) - \mathbf{v}\|_2^2\right].
\end{equation}
At inference time, a video sample is generated by integrating the learned ordinary differential equation (ODE)
\begin{equation}
\frac{d\mathbf{x}_t}{dt} = \mathbf{v}_\theta(\mathbf{x}_t, t),
\end{equation}
starting from a pure Gaussian noise $\mathbf{x}_1 \sim \mathcal{N}(0, \mathbf{I})$ and solving it from $t=1$ to $t=0$ using a numerical solver. 

Building upon these principles, our proposed RCM framework employs a flow-matching-based video diffusion backbone to learn continuous rotational motion in a physically consistent manner, enabling high-fidelity 3D character generation with stable viewpoint transitions.

\subsection{Model Architecture}

The model architecture of RCM is shown in Figure \ref{fig:mode_arch_train_strategies}.
Our RCM is built upon the state-of-the-art video diffusion architecture Wan 2.2~\cite{wan2025wan}, which consists of two complementary components—a low-noise model and a high-noise model—operating over different timesteps to ensure both fine-grained detail and stable motion dynamics.

To enable explicit control over camera viewpoint, we design an external lightweight module termed Camera Encoder, responsible for processing and integrating camera pose information into the generative process. 
Following recent advances in camera-conditioned generation~\cite{he2025cameractrl,li2025hunyuan}, we represent each camera pose using Plücker embeddings~\cite{sitzmann2021light}.
This representation provides a geometrically meaningful way to capture per-pixel viewing information, thereby preserving spatial relationships between image pixels and their corresponding 3D rays.
The extracted Plücker embeddings are processed by Camera Encoder through a downsampling pathway to match the resolution of the input video latent. 
The resulting camera latent is then element-wise summed with the input video latent before being fed into the RCM backbone. 
This conditioning design allows RCM to flexibly generate videos from arbitrary initial camera poses while maintaining consistent motion and viewpoint transitions.
Besides, for image conditioning, to prevent the model from misinterpreting the reference images as initial frames and producing unintended image continuations, we place the latents of the reference images at the end of the video latent sequence, following the approach in \cite{chen2025humo}.

\subsection{Training Strategy Decomposition}
To effectively equip RCM with the abilities of pose canonicalization, high-resolution generation, controllable viewpoints, and multi-view conditioning, we design a progressive training strategy, as shown in Figure \ref{fig:mode_arch_train_strategies}.
Specifically, the training process is divided into three sequential stages, each designed to achieve a distinct learning objective.

\textbf{Stage I: Pose Canonicalization.} 
In this stage, the model is trained to normalize character appearances into a canonical pose. 
Specifically, the input data consist of images depicting characters in diverse and random poses, while the target output is a static video of the same character shown in a canonical A/T pose. 
During this stage, the Camera Encoder module is not incorporated, as camera pose information is unavailable in the corresponding training data of this stage. 
To improve the generalization capability of RCM, we randomly composite each character onto different background images, encouraging the model to focus on reconstructing the character itself rather than the surrounding environment. 
In implementation, the entire model is fine-tuned end-to-end during this optimization stage. 
Through this process, RCM effectively learns to canonicalize character poses, producing stable static videos that depict consistent A/T-pose representations across diverse characters.

\textbf{Stage II: Viewpoint Initialization.} 
Once pose canonicalization has been established, the second stage introduces camera pose conditioning to enable controllable viewpoints in the generated videos. 
The input data at this stage comprises character images with varied poses and backgrounds, accompanied by randomly sampled camera pose parameters. 
The target output remains a static video in which the observation viewpoint changes according to the provided camera pose. 
During this stage, the Camera Encoder module is first trained independently while keeping all other parameters frozen to ensure fast camera embedding learning. 
Subsequently, the entire model is jointly fine-tuned to achieve full integration between appearance and camera conditioning. 
As a result, RCM acquires controllable viewpoint capability, enabling users to synthesize static character videos from arbitrary perspective positions.

\textbf{Stage III: Character Rotation.} 
In the final stage, RCM is trained to synthesize dynamic character rotations covering a complete viewing orbit. 
The input data are identical to those used in Stage II, while the target output is a video depicting the character smoothly rotating in place, conditioned on the provided camera pose information. 
During this stage, the entire model is fine-tuned end-to-end to optimize rotational motion generation. 
As a result, RCM attains the capability to generate videos of characters rotating in their canonical A/T poses, while enabling users to observe the rotation from arbitrary viewpoints.

Besides, during all three training stages, we provide RCM with multiple input character images in random poses and backgrounds (\emph{i.e.}, 1$\sim$4 images). 
This design choice reflects practical user scenarios, where multiple reference images may be uploaded to capture different aspects of the same character, considering that users may want to keep certain parts of their character designs unchanged. 
To further enhance the visual quality and preserve fine-grained details, RCM is trained at a resolution of $1024 \times 1024$.

\section{Experiments}
\label{sec:experiments}
\subsection{Implementation Details}
In our experiments, we adopted Wan 2.2 \cite{wan2025wan} as our base model due to its exceptional video generation quality. 
Considering Wan 2.2 features a Mixture-of-Experts (MoE) architecture, using a low-noise model and a high-noise model targeting different timesteps, we applied our model architecture modifications and training strategies to both the low-noise model and the high-noise model, and trained these models separately according to different timestep values.

\begin{figure}[t]
  \centering
   \includegraphics[width=1\columnwidth]{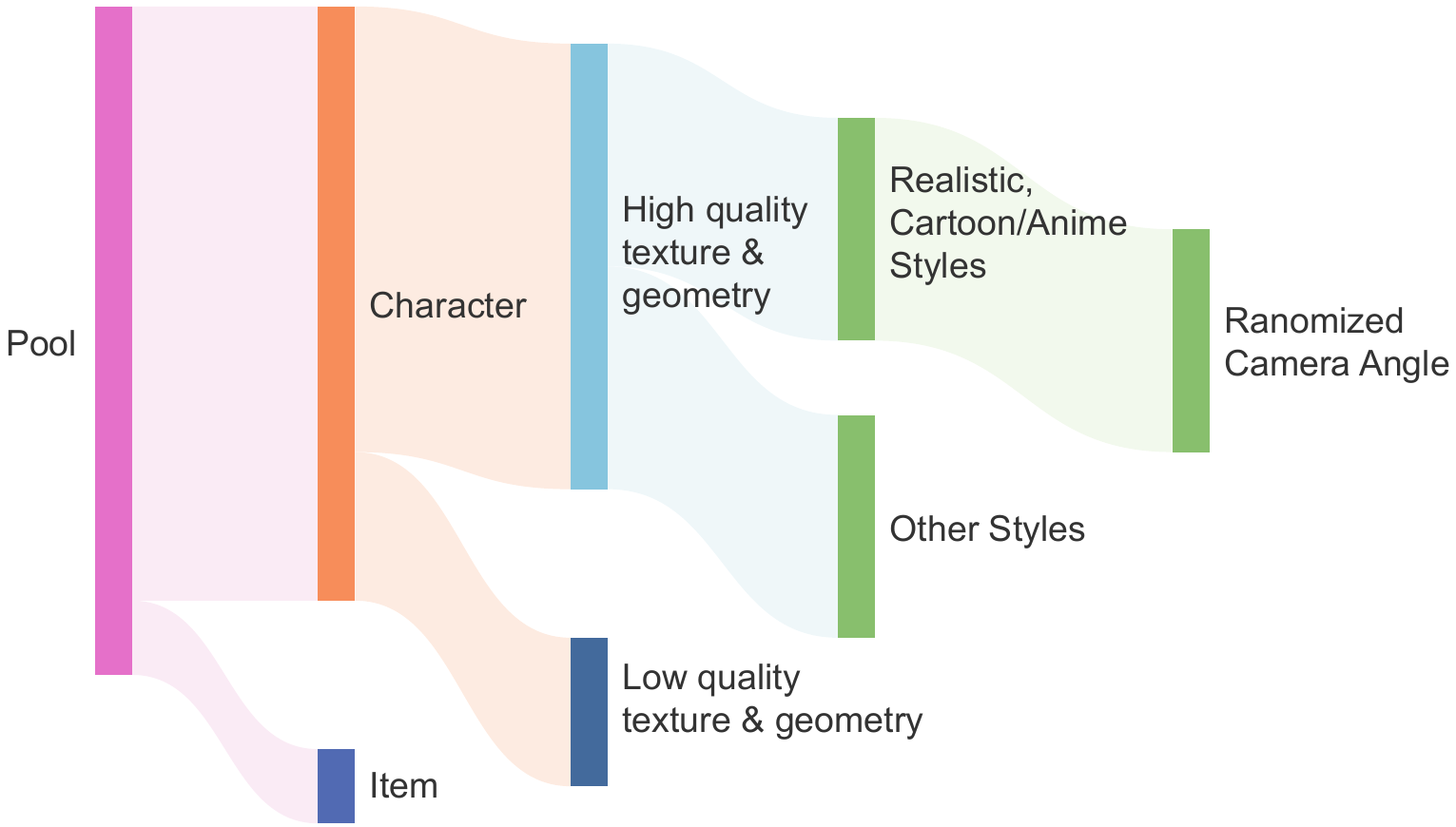}
   \caption{\textbf{Training data filtering pipeline.} We curated our dataset from a general pool, separating character and item data. Characters underwent quality-based filtering, retaining only high-quality geometry and textures. These were further categorized by style (realistic, cartoon/anime, and others) and augmented with randomized camera angles for viewpoint diversity.}
   \label{fig:dataset}
\end{figure}

To train our RCM, we curated and manually rendered approximately 46K proprietary character models, each featuring a standard A/T pose alongside multiple complex poses. For camera configurations, we employed three viewpoints per character: one horizontal observation point and two randomly sampled viewpoints with camera distances ranging from 4 to 7 units and viewing angles spanning $-\frac{\pi}{4}$ to $\frac{\pi}{4}$. 
This process yielded approximately 120K training videos in total after manual filtering. The detailed data filtering pipeline is shown in Figure \ref{fig:dataset}.
\subsection{Evaluation Protocol}
\textbf{Evaluation benchmarks.} Due to the lack of high-quality character animation datasets, we propose two challenging benchmarks encompassing both in-the-wild and complex character cases.
The first benchmark, dubbed \textbf{RCM-Wild}, comprises diverse character images. Specifically, we engage domain experts to design prompts covering multiple dimensions, including gender, art style, hair design, costumes, facial expressions, and poses. These prompts are then fed into state-of-the-art text-to-image generative models \cite{seedream2025seedream,cao2025hunyuanimage,niji2022}, yielding 756 candidate images. Domain experts subsequently curate this collection through rigorous quality filtering, retaining 113 high-quality examples for evaluation.
The second benchmark, termed \textbf{RCM‑Hard}, consists of challenging character images characterized by intricate textures, elaborate accessories, and complex structural details that rigorously test fine‑grained generation capabilities.
It includes 140 high‑quality character models created by domain experts, each containing three images in random poses and one reference video in the standard A pose.
Both benchmarks will be publicly released to advance community research in character generation.

\noindent \textbf{Baselines}. We adopted multi-view diffusion models and image-to-video generative  models as baselines, including CharacterGen \cite{peng2024charactergen}, SyncDreamer \cite{liu2023syncdreamer}, SV3D \cite{voleti2024sv3d}, Epidiff \cite{huang2024epidiff}, Hi3D \cite{yang2024hi3d}, AR-1-to-3 \cite{zhang2025ar}, Wan 2.1 \cite{wan2025wan}, Wan 2.2 \cite{wan2025wan} for both qualitative and quantitative comparisons.
  
\subsection{Qualitative Comparisons}

\begin{figure*}[t]
  \centering
   \includegraphics[width=1\textwidth]{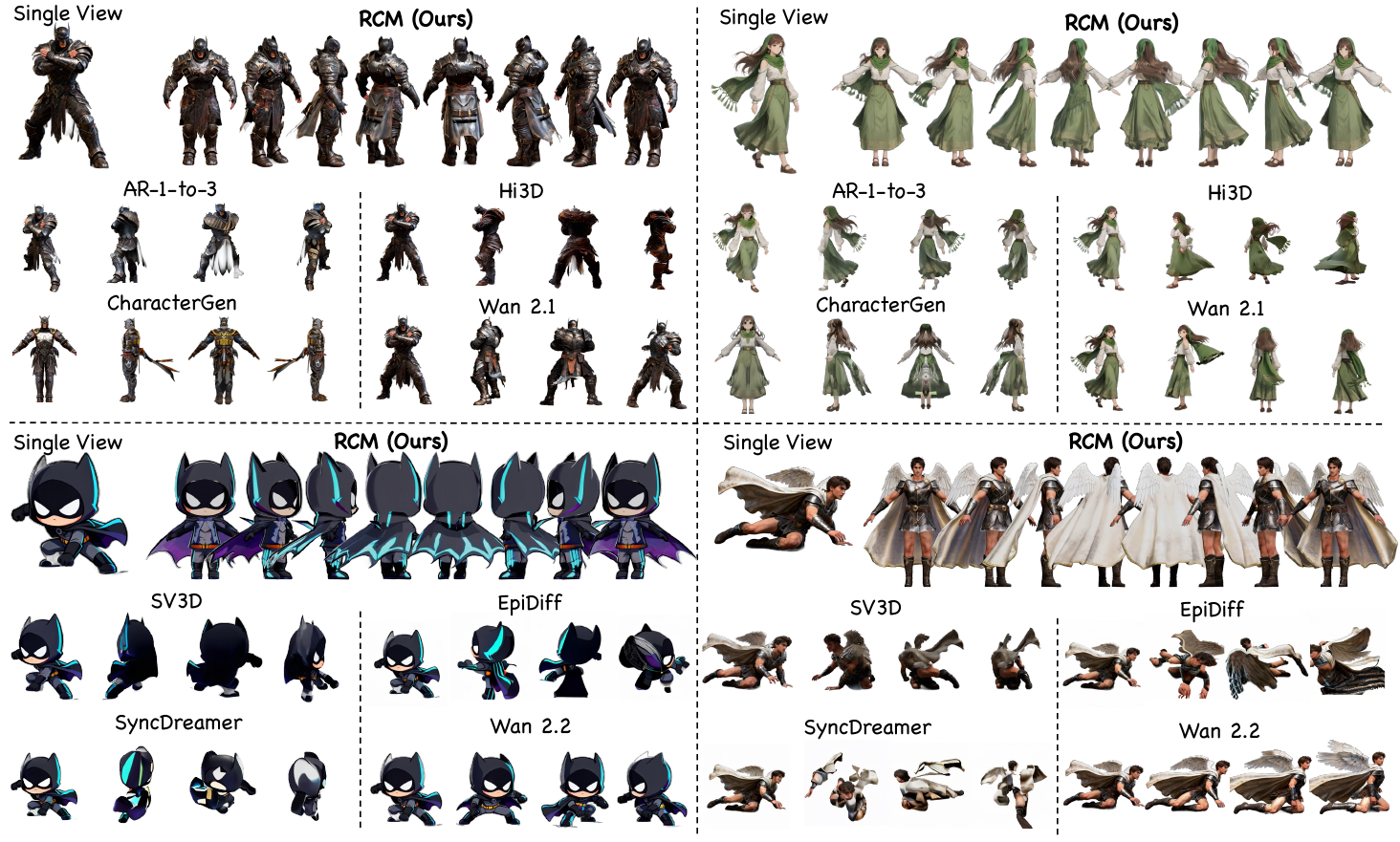}
   \caption{Qualitative comparisons with previous methods, including  CharacterGen \cite{peng2024charactergen}, Hi3D \cite{yang2024hi3d}, AR-1-to-3 \cite{zhang2025ar}, Wan 2.1 \cite{wan2025wan}, SyncDreamer \cite{liu2023syncdreamer}, SV3D \cite{voleti2024sv3d}, Epidiff \cite{huang2024epidiff} and Wan 2.2 \cite{wan2025wan}. Our method effectively transformed characters with complex poses into standard A/T poses while preserving a complete orbital visualization of each character.}
   \label{fig:single-view-2}
\end{figure*}

We conducted qualitative comparisons on our proposed RCM‑Wild benchmark, as illustrated in Figure \ref{fig:single-view-2}.
For Wan 2.1 \cite{wan2025wan} and Wan 2.2 \cite{wan2025wan}, we prompted the models to generate videos depicting a full orbit of each input character.
As shown in these figures, most multi‑view diffusion models failed to canonicalize characters into the standard A/T pose.
Although CharacterGen \cite{peng2024charactergen} successfully performed the pose transformation, it often struggled to preserve character identity when processing images with complex poses and fine‑grained details.
In contrast, the video diffusion models Wan 2.1/2.2 demonstrated stronger identity preservation but frequently failed to generate canonical A/T poses or complete orbital trajectories.
Our proposed method achieved both accurate pose canonicalization and consistent identity preservation—even with 2D‑based character inputs—while simultaneously producing smooth and complete orbital visualizations.

\subsection{Quantitative Comparisons}

\begin{table}[t]
\centering
\caption{Quantitative comparisons in the user study on the RCM-Wild benchmark. The results demonstrate that our method delivered superior performance in terms of image-video consistency (IVC), prompt-video consistency (PVC), aesthetic quality (AQ), motion quality (MQ) and pose canonicalization (PC).}
{
\linespread{1.0}
\setlength\tabcolsep{5pt}
\small
{%
\begin{tabular}{lcccccc}
\toprule
\textbf{Method} & \textbf{IVC} & \textbf{PVC} & \textbf{AQ} & \textbf{MQ}& \textbf{PC}& \textbf{Overall} \\
\midrule
SyncDreamer& 1.28 & 1.31 & 1.10 & 1.16  & 1.12 & 1.15 \\
Epidiff & 1.42 & 1.40 & 1.12 & 1.16 & 1.17  & 1.19 \\
Hi3D & 2.00 & 1.91 & 1.33 & 1.72  & 1.48& 1.56 \\
AR-1-to-3  & 2.07 & 2.09 & 1.72 & 1.92  & 1.62 & 1.80 \\
CharacterGen & 2.52 & 2.97 & 2.51 & 2.79 & 2.48 & 2.64  \\
Wan 2.1 & 2.86 & 2.09 & 2.51 & 1.78 & 1.54 & 2.05  \\
Wan 2.2 & 3.73 & 2.92 & 3.56& 3.05  & 2.26 & 3.03\\
\midrule
\textbf{RCM(Ours)} & \textbf{3.73} & \textbf{4.40} & \textbf{4.15} & \textbf{4.33}  & \textbf{4.36}&\textbf{4.25} \\
\bottomrule
\end{tabular}
}
}
\label{tab:user_study}
\end{table}

\noindent \textbf{Results on the RCM-Wild benchmark}. We conducted a user study on the proposed RCM-Wild benchmark to evaluate perceptual quality and subjective preferences.
Specifically, the evaluation was based on five criteria designed to capture different aspects of generation quality: (1) subject consistency between the input character image and the output video (Image-Video Consistency), measuring how well the generated video preserves the character's distinctive features and identity; (2) consistency of the video with the input prompt (Prompt-Video Consistency), assessing alignment between the text description and visual content; (3) video aesthetic quality, evaluating overall visual appeal, color harmony, and rendering quality; (4) motion quality, measuring the smoothness and naturalness of viewpoint transitions; and (5) pose canonicalization, assessing the accuracy and completeness of transformation to standard A/T pose.
Each criterion was rated on a 1–5 Likert scale, where 1 represented the lowest quality (poor/inconsistent) and 5 represented the highest quality (excellent/highly consistent).
A total of 18 domain experts with professional experience in computer graphics, computer vision, or 3D character design participated in the study to assess the generated character videos according to these criteria.
Participants were shown randomly ordered characters from all methods for the same input to minimize bias.
As shown in Table \ref{tab:user_study}, experimental results demonstrate that our method significantly outperformed existing approaches across all five evaluation perspectives, achieving the highest average ratings in image-video consistency, prompt-video consistency, aesthetic quality, motion quality, and pose canonicalization.
The superior performance in image-video consistency validated our method's ability to preserve character identity through complex pose transformations, while the high scores in motion quality and pose canonicalization confirmed the effectiveness of our progressive training strategy and geometry-aware camera conditioning.
These human evaluation results provided perceptual validation that RCM generated character videos with higher subjective quality as perceived by expert evaluators.

\noindent \textbf{Results on the RCM-Hard benchmark}.
We performed quantitative comparisons on the proposed RCM-Hard benchmark by reporting PSNR and SSIM metrics for novel view synthesis, following the evaluation protocols established in previous studies \cite{liu2023syncdreamer,yang2024hi3d,peng2024charactergen}. 
Specifically, we used the same single-view image as input for each character across all methods to ensure fair comparison.
Results are summarized in Table \ref{tab:reconstruction_quality}, demonstrating that our method generated images that are semantically aligned with the input while maintaining superior multi-view consistency in both color and geometry, demonstrating the comprehensive advantages of our approach.

\begin{table}[t]
\centering
\caption{Quantitative comparison on the RCM-Hard benchmark for novel view synthesis. We evaluated our method against state-of-the-art approaches across three metrics: PSNR and SSIM. Results demonstrate that RCM achieved superior performance across all metrics, outperforming both multi-view generation methods and video diffusion models baselines.}
{
\linespread{1.0}
\setlength\tabcolsep{10pt}
{%
\begin{tabular}{lccc}
\toprule
\textbf{Method} & \textbf{PSNR}\,$\uparrow$ & \textbf{SSIM}\,$\uparrow$   \\
\midrule
SyncDreamer & 11.90 & 0.87  \\
SV3D & 11.84 & 0.87 \\
Epidiff & 11.80 & 0.87 \\
Hi3D & 12.10 & 0.88  \\
AR-1-to-3 & 11.62 & 0.85 \\
CharacterGen & 9.31 & 0.80  \\
Wan 2.1 & 7.33 & 0.64\\ 
Wan 2.2 & 16.78	& 0.84   \\ 
\textbf{RCM(Ours)} & \textbf{19.41} & \textbf{0.89} \\ 
\bottomrule
\end{tabular}
}
}
\label{tab:reconstruction_quality}
\end{table}

\subsection{Ablation Studies}

\begin{figure}[t]
  \centering
   \includegraphics[width=1\columnwidth]{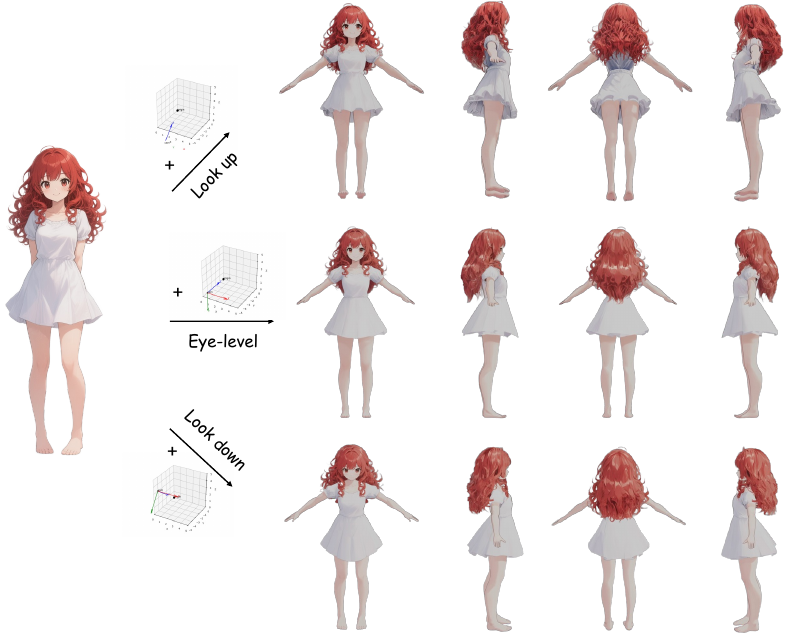}
   \caption{Effects of different camera pose conditions. Users can specify arbitrary viewing angles—including top‑down, bottom‑up, and eye-level perspectives—with our method reliably generating geometrically accurate results that retain character appearance.}
   \label{fig:camerapose}
\end{figure}
\textbf{Camera Pose Conditions}. 
In this section, we demonstrate the effect of camera pose conditioning in our proposed RCM.
As shown in Figure \ref{fig:camerapose}, the model produced consistent results across different camera poses—including top-down, bottom-up, and frontal views—accurately reflecting the intended viewpoints while preserving the character's structure and identity.
These capabilities allow users to freely observe characters from arbitrary viewpoints, enabling flexible and geometry-aware visual exploration.

\noindent \textbf{Multi-view Condition Images}.
\begin{figure}[t]
  \centering
   \includegraphics[width=0.5\textwidth]{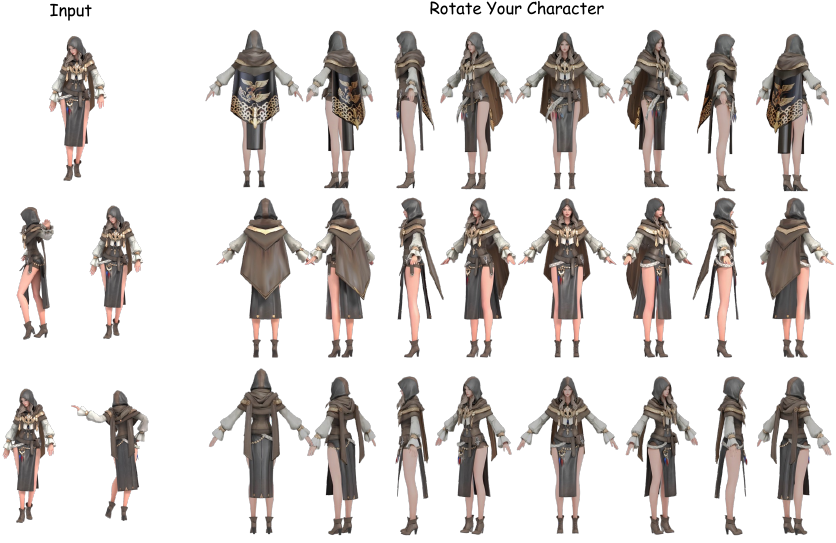}
   \caption{Effects of multi-view images. Single-image input produced plausible completion of occluded regions (\textit{i.e.}, the first row); additional viewpoint images enabled geometric refinement through cross-view consensus (\textit{i.e.}, the second and the third rows). Note that in the first row, we prompted RCM to generate a feather pattern on the character's back.}
   \label{fig:multiview}
\end{figure}
In this section, we demonstrate the effect of providing multi‑view images as inputs to our proposed RCM.
Specifically, as shown in Figure \ref{fig:multiview}, when given a single‑view image, RCM can reasonably infer occluded or missing parts, producing a complete and coherent reconstruction of the character. 
When multiple views were provided, RCM effectively leveraged the additional information to align details across views, resulting in more accurate geometry, consistent appearance, and higher overall fidelity.
This capability provides users with a flexible interface for character creation, enabling them to customize and refine their designs according to their preferences.

\begin{figure}[t]
  \centering
   \includegraphics[width=0.5\textwidth]{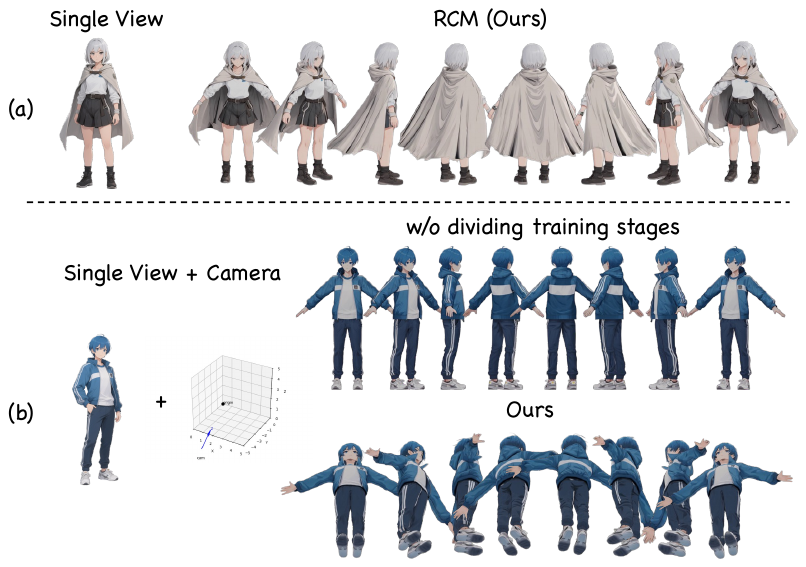}
   \caption{Generalization and training stages. (a) Our method generalized to Wan 2.1, generating consistent orbital character views. (b) Training without stage division caused the model to ignore camera pose conditions (top), as pose canonicalization and viewpoint control become entangled. Our progressive strategy (bottom) successfully disentangled these goals, achieving precise camera-conditioned generation.}
   \label{fig:ablation_wan21}
\end{figure}

\noindent \textbf{Generalization to Other Base Models}. 
In this section, we applied our design to another video diffusion model, Wan 2.1 \cite{wan2025wan}, to demonstrate the generalizability of our proposed method.
As shown in Figure \ref{fig:ablation_wan21} (a), the adapted model successfully generated the coherent multi-view character video from a single-view image, maintaining consistent appearance and geometry throughout the orbital rotation.
This experiment validates that our progressive training strategy and Camera Encoder design are not tightly coupled to a specific backbone architecture, but can be effectively transferred across different video diffusion models.

\noindent \textbf{Effects of Training Stages}. In this section, we examined the impact of our proposed progressive training strategy.
Specifically, we conducted an ablation study in which RCM was trained directly without dividing the process into separate stages.
In this setting, the newly introduced Camera Encoder and the video diffusion backbone were optimized jointly to learn pose canonicalization, viewpoint control, and character rotation simultaneously.
However, as shown in Figure \ref{fig:ablation_wan21} (b), this joint training led to suboptimal performance—particularly in controllable viewpoint generation—highlighting the necessity of the staged training design.
This failure mode occurred because the model struggled to disentangle these objectives, resulting in degraded camera controllability where the generated views did not accurately reflect the input camera parameters.

\begin{figure}[t]
  \centering
   \includegraphics[width=1\columnwidth]{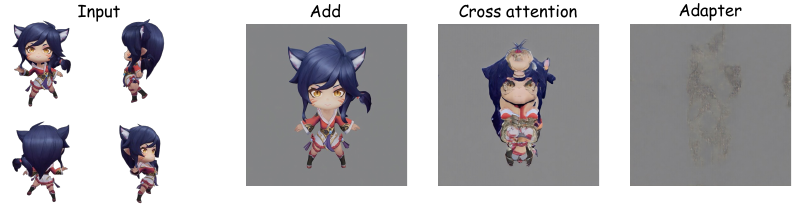}
   \caption{\textbf{Ablation on camera pose injection methods.} We compared different strategies for incorporating camera pose information: direct addition to latent features (Add), cross-attention mechanisms, and adapter-based injection (IP-Adapter style). Addition preserved the pretrained model's knowledge, while cross-attention and adapter methods caused visual artifacts and quality degradation, indicating significantly longer training.}
   \label{fig:camera_injection}
\end{figure}

\noindent \textbf{Different Methods for Injecting Camera Poses}. In this section, we investigated alternative strategies for integrating camera pose information into our proposed RCM.
Specifically, we experimented with injecting camera pose features through cross-attention mechanisms and the approach introduced in IP-Adapter \cite{ye2023ip}.
However, as shown in Figure \ref{fig:camera_injection}, these methods disrupted the pretrained knowledge of the underlying video diffusion model, which led to severe visual distortions and degraded video quality, indicating a need for substantially longer training to achieve convergence.

\section{Conclusion}
\label{sec:conclusion}
In this paper, we have presented RCM (Rotate your Character Model), an image-to-video diffusion framework for high-quality character generation.
Through a progressive training strategy and a geometry-aware Camera Encoder, RCM has achieved accurate pose canonicalization, high-resolution generations, controllable viewpoints, multi-view conditioning, along with superior identity preservation.
Extensive experiments across challenging benchmarks have demonstrated its significant advantages over existing multi-view and video diffusion methods, highlighting its potential for controllable 3D character creation.
We believe RCM represents an important step toward democratizing 3D content creation, making sophisticated character visualization accessible to broader creative communities.
Future work will explore extending RCM to handle dynamic character animations and integrating it with downstream 3D reconstruction pipelines for fully automated asset generation.

{
    \small
    \bibliographystyle{ieeenat_fullname}
    \bibliography{main}

@String(TOG= {ACM Trans. Graph.})

@String(TOG   = {ACM TOG})

@article{peng2024charactergen,
  title={Charactergen: Efficient 3d character generation from single images with multi-view pose canonicalization},
  author={Peng, Hao-Yang and Zhang, Jia-Peng and Guo, Meng-Hao and Cao, Yan-Pei and Hu, Shi-Min},
  journal={ACM Transactions on Graphics (TOG)},
  volume={43},
  number={4},
  pages={1--13},
  year={2024},
  publisher={ACM New York, NY, USA}
}

@article{shi2023zero123++,
  title={Zero123++: a single image to consistent multi-view diffusion base model},
  author={Shi, Ruoxi and Chen, Hansheng and Zhang, Zhuoyang and Liu, Minghua and Xu, Chao and Wei, Xinyue and Chen, Linghao and Zeng, Chong and Su, Hao},
  journal={arXiv preprint arXiv:2310.15110},
  year={2023}
}

@inproceedings{liu2023zero,
  title={Zero-1-to-3: Zero-shot one image to 3d object},
  author={Liu, Ruoshi and Wu, Rundi and Van Hoorick, Basile and Tokmakov, Pavel and Zakharov, Sergey and Vondrick, Carl},
  booktitle={Proceedings of the IEEE/CVF international conference on computer vision},
  pages={9298--9309},
  year={2023}
}

@inproceedings{he2025cameractrl,
  title={Cameractrl: Enabling camera control for video diffusion models},
  author={He, Hao and Xu, Yinghao and Guo, Yuwei and Wetzstein, Gordon and Dai, Bo and Li, Hongsheng and Yang, Ceyuan},
  booktitle={The Thirteenth International Conference on Learning Representations},
  year={2025}
}

@article{poole2022dreamfusion,
  title={Dreamfusion: Text-to-3d using 2d diffusion},
  author={Poole, Ben and Jain, Ajay and Barron, Jonathan T and Mildenhall, Ben},
  journal={arXiv preprint arXiv:2209.14988},
  year={2022}
}

@inproceedings{voleti2024sv3d,
  title={Sv3d: Novel multi-view synthesis and 3d generation from a single image using latent video diffusion},
  author={Voleti, Vikram and Yao, Chun-Han and Boss, Mark and Letts, Adam and Pankratz, David and Tochilkin, Dmitry and Laforte, Christian and Rombach, Robin and Jampani, Varun},
  booktitle={European Conference on Computer Vision},
  pages={439--457},
  year={2024},
  organization={Springer}
}

@article{chen2024v3d,
  title={V3d: Video diffusion models are effective 3d generators},
  author={Chen, Zilong and Wang, Yikai and Wang, Feng and Wang, Zhengyi and Liu, Huaping},
  journal={arXiv preprint arXiv:2403.06738},
  year={2024}
}

@article{melas20243d,
  title={Im-3d: Iterative multiview diffusion and reconstruction for high-quality 3d generation},
  author={Melas-Kyriazi, Luke and Laina, Iro and Rupprecht, Christian and Neverova, Natalia and Vedaldi, Andrea and Gafni, Oran and Kokkinos, Filippos},
  journal={arXiv preprint arXiv:2402.08682},
  year={2024}
}

@article{zuo2024videomv,
  title={Videomv: Consistent multi-view generation based on large video generative model},
  author={Zuo, Qi and Gu, Xiaodong and Qiu, Lingteng and Dong, Yuan and Yuan, Weihao and Peng, Rui and Zhu, Siyu and Bo, Liefeng and Dong, Zilong and Huang, Qixing and others}
}

@inproceedings{yang2024hi3d,
  title={Hi3d: Pursuing high-resolution image-to-3d generation with video diffusion models},
  author={Yang, Haibo and Chen, Yang and Pan, Yingwei and Yao, Ting and Chen, Zhineng and Ngo, Chong-Wah and Mei, Tao},
  booktitle={Proceedings of the 32nd ACM International Conference on Multimedia},
  pages={6870--6879},
  year={2024}
}

@inproceedings{xingmirror,
  title={MIRROR: Make Your Object-Level Multi-View Generation More Consistent with Training-Free Rectification},
  author={Xing, Tianchi and Li, Bonan and Han, Congying and Qiu, Xinmin and Zhang, Zicheng and Guo, Tiande},
  booktitle={Forty-second International Conference on Machine Learning}
}

@article{liu2023syncdreamer,
  title={Syncdreamer: Generating multiview-consistent images from a single-view image},
  author={Liu, Yuan and Lin, Cheng and Zeng, Zijiao and Long, Xiaoxiao and Liu, Lingjie and Komura, Taku and Wang, Wenping},
  journal={arXiv preprint arXiv:2309.03453},
  year={2023}
}

@inproceedings{lin2023magic3d,
  title={Magic3d: High-resolution text-to-3d content creation},
  author={Lin, Chen-Hsuan and Gao, Jun and Tang, Luming and Takikawa, Towaki and Zeng, Xiaohui and Huang, Xun and Kreis, Karsten and Fidler, Sanja and Liu, Ming-Yu and Lin, Tsung-Yi},
  booktitle={Proceedings of the IEEE/CVF conference on computer vision and pattern recognition},
  pages={300--309},
  year={2023}
}

@inproceedings{esser2024scaling,
  title={Scaling rectified flow transformers for high-resolution image synthesis},
  author={Esser, Patrick and Kulal, Sumith and Blattmann, Andreas and Entezari, Rahim and M{\"u}ller, Jonas and Saini, Harry and Levi, Yam and Lorenz, Dominik and Sauer, Axel and Boesel, Frederic and others},
  booktitle={Forty-first international conference on machine learning},
  year={2024}
}

@inproceedings{rombach2022high,
  title={High-resolution image synthesis with latent diffusion models},
  author={Rombach, Robin and Blattmann, Andreas and Lorenz, Dominik and Esser, Patrick and Ommer, Bj{\"o}rn},
  booktitle={Proceedings of the IEEE/CVF conference on computer vision and pattern recognition},
  pages={10684--10695},
  year={2022}
}

@misc{labs2025flux1kontextflowmatching,
      title={FLUX.1 Kontext: Flow Matching for In-Context Image Generation and Editing in Latent Space},
      author={Black Forest Labs and Stephen Batifol and Andreas Blattmann and Frederic Boesel and Saksham Consul and Cyril Diagne and Tim Dockhorn and Jack English and Zion English and Patrick Esser and Sumith Kulal and Kyle Lacey and Yam Levi and Cheng Li and Dominik Lorenz and Jonas Müller and Dustin Podell and Robin Rombach and Harry Saini and Axel Sauer and Luke Smith},
      year={2025},
      eprint={2506.15742},
      archivePrefix={arXiv},
      primaryClass={cs.GR},
      url={https://arxiv.org/abs/2506.15742},
}

@misc{flux2024,
    author={Black Forest Labs},
    title={FLUX},
    year={2024}
}

@inproceedings{chen2023fantasia3d,
  title={Fantasia3d: Disentangling geometry and appearance for high-quality text-to-3d content creation},
  author={Chen, Rui and Chen, Yongwei and Jiao, Ningxin and Jia, Kui},
  booktitle={Proceedings of the IEEE/CVF international conference on computer vision},
  pages={22246--22256},
  year={2023}
}

@article{wang2023prolificdreamer,
  title={Prolificdreamer: High-fidelity and diverse text-to-3d generation with variational score distillation},
  author={Wang, Zhengyi and Lu, Cheng and Wang, Yikai and Bao, Fan and Li, Chongxuan and Su, Hang and Zhu, Jun},
  journal={Advances in neural information processing systems},
  volume={36},
  pages={8406--8441},
  year={2023}
}

@inproceedings{zhuhifa,
  title={HIFA: High-fidelity Text-to-3D Generation with Advanced Diffusion Guidance},
  author={Zhu, Junzhe and Zhuang, Peiye and Koyejo, Sanmi},
  booktitle={The Twelfth International Conference on Learning Representations}
}

@inproceedings{huangdreamtime,
  title={DreamTime: An Improved Optimization Strategy for Diffusion-Guided 3D Generation},
  author={Huang, Yukun and Wang, Jianan and Shi, Yukai and Tang, Boshi and Qi, Xianbiao and Zhang, Lei},
  booktitle={The Twelfth International Conference on Learning Representations}
}

@inproceedings{tangdreamgaussian,
  title={DreamGaussian: Generative Gaussian Splatting for Efficient 3D Content Creation},
  author={Tang, Jiaxiang and Ren, Jiawei and Zhou, Hang and Liu, Ziwei and Zeng, Gang},
  booktitle={The Twelfth International Conference on Learning Representations}
}

@inproceedings{yi2024gaussiandreamer,
  title={Gaussiandreamer: Fast generation from text to 3d gaussians by bridging 2d and 3d diffusion models},
  author={Yi, Taoran and Fang, Jiemin and Wang, Junjie and Wu, Guanjun and Xie, Lingxi and Zhang, Xiaopeng and Liu, Wenyu and Tian, Qi and Wang, Xinggang},
  booktitle={Proceedings of the IEEE/CVF Conference on Computer Vision and Pattern Recognition},
  pages={6796--6807},
  year={2024}
}

@inproceedings{shimvdream,
  title={MVDream: Multi-view Diffusion for 3D Generation},
  author={Shi, Yichun and Wang, Peng and Ye, Jianglong and Mai, Long and Li, Kejie and Yang, Xiao},
  booktitle={The Twelfth International Conference on Learning Representations}
}

@inproceedings{long2024wonder3d,
  title={Wonder3d: Single image to 3d using cross-domain diffusion},
  author={Long, Xiaoxiao and Guo, Yuan-Chen and Lin, Cheng and Liu, Yuan and Dou, Zhiyang and Liu, Lingjie and Ma, Yuexin and Zhang, Song-Hai and Habermann, Marc and Theobalt, Christian and others},
  booktitle={Proceedings of the IEEE/CVF conference on computer vision and pattern recognition},
  pages={9970--9980},
  year={2024}
}

@article{liu2023one,
  title={One-2-3-45: Any single image to 3d mesh in 45 seconds without per-shape optimization},
  author={Liu, Minghua and Xu, Chao and Jin, Haian and Chen, Linghao and Varma T, Mukund and Xu, Zexiang and Su, Hao},
  journal={Advances in Neural Information Processing Systems},
  volume={36},
  pages={22226--22246},
  year={2023}
}

@article{wang2023imagedream,
  title={Imagedream: Image-prompt multi-view diffusion for 3d generation},
  author={Wang, Peng and Shi, Yichun},
  journal={arXiv preprint arXiv:2312.02201},
  year={2023}
}

@inproceedings{liinstant3d,
  title={Instant3D: Fast Text-to-3D with Sparse-view Generation and Large Reconstruction Model},
  author={Li, Jiahao and Tan, Hao and Zhang, Kai and Xu, Zexiang and Luan, Fujun and Xu, Yinghao and Hong, Yicong and Sunkavalli, Kalyan and Shakhnarovich, Greg and Bi, Sai},
  booktitle={The Twelfth International Conference on Learning Representations}
}

@inproceedings{liu2024one,
  title={One-2-3-45++: Fast single image to 3d objects with consistent multi-view generation and 3d diffusion},
  author={Liu, Minghua and Shi, Ruoxi and Chen, Linghao and Zhang, Zhuoyang and Xu, Chao and Wei, Xinyue and Chen, Hansheng and Zeng, Chong and Gu, Jiayuan and Su, Hao},
  booktitle={Proceedings of the IEEE/CVF conference on computer vision and pattern recognition},
  pages={10072--10083},
  year={2024}
}

@inproceedings{tang2024lgm,
  title={Lgm: Large multi-view gaussian model for high-resolution 3d content creation},
  author={Tang, Jiaxiang and Chen, Zhaoxi and Chen, Xiaokang and Wang, Tengfei and Zeng, Gang and Liu, Ziwei},
  booktitle={European Conference on Computer Vision},
  pages={1--18},
  year={2024},
  organization={Springer}
}

@article{xu2024instantmesh,
  title={Instantmesh: Efficient 3d mesh generation from a single image with sparse-view large reconstruction models},
  author={Xu, Jiale and Cheng, Weihao and Gao, Yiming and Wang, Xintao and Gao, Shenghua and Shan, Ying},
  journal={arXiv preprint arXiv:2404.07191},
  year={2024}
}

@inproceedings{lin2024consistent123,
  title={Consistent123: One image to highly consistent 3d asset using case-aware diffusion priors},
  author={Lin, Yukang and Han, Haonan and Gong, Chaoqun and Xu, Zunnan and Zhang, Yachao and Li, Xiu},
  booktitle={Proceedings of the 32nd ACM International Conference on Multimedia},
  pages={6715--6724},
  year={2024}
}

@inproceedings{hu2024mvd,
  title={Mvd-fusion: Single-view 3d via depth-consistent multi-view generation},
  author={Hu, Hanzhe and Zhou, Zhizhuo and Jampani, Varun and Tulsiani, Shubham},
  booktitle={Proceedings of the IEEE/CVF Conference on Computer Vision and Pattern Recognition},
  pages={9698--9707},
  year={2024}
}

@article{tang2023MVDiffusion,
  title={MVDiffusion: Enabling Holistic Multi-view Image Generation with Correspondence-Aware Diffusion},
  author={Tang, Shitao and Zhang, Fuayng and Chen, Jiacheng and Wang, Peng and Yasutaka, Furukawa},
  journal={arXiv preprint 2307.01097},
  year={2023}
}

@article{yang2024cogvideox,
  title={Cogvideox: Text-to-video diffusion models with an expert transformer},
  author={Yang, Zhuoyi and Teng, Jiayan and Zheng, Wendi and Ding, Ming and Huang, Shiyu and Xu, Jiazheng and Yang, Yuanming and Hong, Wenyi and Zhang, Xiaohan and Feng, Guanyu and others},
  journal={arXiv preprint arXiv:2408.06072},
  year={2024}
}

@article{kong2024hunyuanvideo,
  title={Hunyuanvideo: A systematic framework for large video generative models},
  author={Kong, Weijie and Tian, Qi and Zhang, Zijian and Min, Rox and Dai, Zuozhuo and Zhou, Jin and Xiong, Jiangfeng and Li, Xin and Wu, Bo and Zhang, Jianwei and others},
  journal={arXiv preprint arXiv:2412.03603},
  year={2024}
}

@article{wan2025wan,
  title={Wan: Open and advanced large-scale video generative models},
  author={Wan, Team and Wang, Ang and Ai, Baole and Wen, Bin and Mao, Chaojie and Xie, Chen-Wei and Chen, Di and Yu, Feiwu and Zhao, Haiming and Yang, Jianxiao and others},
  journal={arXiv preprint arXiv:2503.20314},
  year={2025}
}

@article{zheng2024open,
  title={Open-sora: Democratizing efficient video production for all},
  author={Zheng, Zangwei and Peng, Xiangyu and Yang, Tianji and Shen, Chenhui and Li, Shenggui and Liu, Hongxin and Zhou, Yukun and Li, Tianyi and You, Yang},
  journal={arXiv preprint arXiv:2412.20404},
  year={2024}
}

@article{lin2024open,
  title={Open-sora plan: Open-source large video generation model},
  author={Lin, Bin and Ge, Yunyang and Cheng, Xinhua and Li, Zongjian and Zhu, Bin and Wang, Shaodong and He, Xianyi and Ye, Yang and Yuan, Shenghai and Chen, Liuhan and others},
  journal={arXiv preprint arXiv:2412.00131},
  year={2024}
}

@article{li2025hunyuan,
  title={Hunyuan-GameCraft: High-dynamic Interactive Game Video Generation with Hybrid History Condition},
  author={Li, Jiaqi and Tang, Junshu and Xu, Zhiyong and Wu, Longhuang and Zhou, Yuan and Shao, Shuai and Yu, Tianbao and Cao, Zhiguo and Lu, Qinglin},
  journal={arXiv preprint arXiv:2506.17201},
  year={2025}
}

@article{sitzmann2021light,
  title={Light field networks: Neural scene representations with single-evaluation rendering},
  author={Sitzmann, Vincent and Rezchikov, Semon and Freeman, Bill and Tenenbaum, Josh and Durand, Fredo},
  journal={Advances in Neural Information Processing Systems},
  volume={34},
  pages={19313--19325},
  year={2021}
}

@article{ho2020denoising,
  title={Denoising diffusion probabilistic models},
  author={Ho, Jonathan and Jain, Ajay and Abbeel, Pieter},
  journal={Advances in neural information processing systems},
  volume={33},
  pages={6840--6851},
  year={2020}
}

@article{dhariwal2021diffusion,
  title={Diffusion models beat gans on image synthesis},
  author={Dhariwal, Prafulla and Nichol, Alexander},
  journal={Advances in neural information processing systems},
  volume={34},
  pages={8780--8794},
  year={2021}
}

@article{blattmann2023stable,
  title={Stable video diffusion: Scaling latent video diffusion models to large datasets},
  author={Blattmann, Andreas and Dockhorn, Tim and Kulal, Sumith and Mendelevitch, Daniel and Kilian, Maciej and Lorenz, Dominik and Levi, Yam and English, Zion and Voleti, Vikram and Letts, Adam and others},
  journal={arXiv preprint arXiv:2311.15127},
  year={2023}
}

@article{ho2022video,
  title={Video diffusion models},
  author={Ho, Jonathan and Salimans, Tim and Gritsenko, Alexey and Chan, William and Norouzi, Mohammad and Fleet, David J},
  journal={Advances in neural information processing systems},
  volume={35},
  pages={8633--8646},
  year={2022}
}

@inproceedings{blattmann2023align,
  title={Align your latents: High-resolution video synthesis with latent diffusion models},
  author={Blattmann, Andreas and Rombach, Robin and Ling, Huan and Dockhorn, Tim and Kim, Seung Wook and Fidler, Sanja and Kreis, Karsten},
  booktitle={Proceedings of the IEEE/CVF conference on computer vision and pattern recognition},
  pages={22563--22575},
  year={2023}
}

@misc{openai2024sora,
  author       = {OpenAI},
  title        = {Sora: Creating video from text},
  year         = {2024},
  note         = {Accessed: 2025-11-07}
}

@inproceedings{chen2024videocrafter2,
  title={Videocrafter2: Overcoming data limitations for high-quality video diffusion models},
  author={Chen, Haoxin and Zhang, Yong and Cun, Xiaodong and Xia, Menghan and Wang, Xintao and Weng, Chao and Shan, Ying},
  booktitle={Proceedings of the IEEE/CVF Conference on Computer Vision and Pattern Recognition},
  pages={7310--7320},
  year={2024}
}

@article{wang2024vidprom,
  title={Vidprom: A million-scale real prompt-gallery dataset for text-to-video diffusion models},
  author={Wang, Wenhao and Yang, Yi},
  journal={Advances in Neural Information Processing Systems},
  volume={37},
  pages={65618--65642},
  year={2024}
}

@inproceedings{chen2024panda,
  title={Panda-70m: Captioning 70m videos with multiple cross-modality teachers},
  author={Chen, Tsai-Shien and Siarohin, Aliaksandr and Menapace, Willi and Deyneka, Ekaterina and Chao, Hsiang-wei and Jeon, Byung Eun and Fang, Yuwei and Lee, Hsin-Ying and Ren, Jian and Yang, Ming-Hsuan and others},
  booktitle={Proceedings of the IEEE/CVF Conference on Computer Vision and Pattern Recognition},
  pages={13320--13331},
  year={2024}
}

@inproceedings{lipmanflow,
  title={Flow Matching for Generative Modeling},
  author={Lipman, Yaron and Chen, Ricky TQ and Ben-Hamu, Heli and Nickel, Maximilian and Le, Matthew},
  booktitle={The Eleventh International Conference on Learning Representations}
}

@inproceedings{liuflow,
  title={Flow Straight and Fast: Learning to Generate and Transfer Data with Rectified Flow},
  author={Liu, Xingchao and Gong, Chengyue and others},
  booktitle={The Eleventh International Conference on Learning Representations}
}

@misc{google2024veo,
  author       = {Google DeepMind},
  title        = {Veo3: Our state-of-the-art video generation model},
  year         = {2025},
  note         = {Accessed: 2025-11-07}
}

@article{gao2025seedance,
  title={Seedance 1.0: Exploring the Boundaries of Video Generation Models},
  author={Gao, Yu and Guo, Haoyuan and Hoang, Tuyen and Huang, Weilin and Jiang, Lu and Kong, Fangyuan and Li, Huixia and Li, Jiashi and Li, Liang and Li, Xiaojie and others},
  journal={arXiv preprint arXiv:2506.09113},
  year={2025}
}

@article{wiedemer2025video,
  title={Video models are zero-shot learners and reasoners},
  author={Wiedemer, Thadd{\"a}us and Li, Yuxuan and Vicol, Paul and Gu, Shixiang Shane and Matarese, Nick and Swersky, Kevin and Kim, Been and Jaini, Priyank and Geirhos, Robert},
  journal={arXiv preprint arXiv:2509.20328},
  year={2025}
}

@article{seedream2025seedream,
  title={Seedream 4.0: Toward next-generation multimodal image generation},
  author={Seedream, Team and Chen, Yunpeng and Gao, Yu and Gong, Lixue and Guo, Meng and Guo, Qiushan and Guo, Zhiyao and Hou, Xiaoxia and Huang, Weilin and Huang, Yixuan and others},
  journal={arXiv preprint arXiv:2509.20427},
  year={2025}
}

@article{cao2025hunyuanimage,
  title={Hunyuanimage 3.0 technical report},
  author={Cao, Siyu and Chen, Hangting and Chen, Peng and Cheng, Yiji and Cui, Yutao and Deng, Xinchi and Dong, Ying and Gong, Kipper and Gu, Tianpeng and Gu, Xiusen and others},
  journal={arXiv preprint arXiv:2509.23951},
  year={2025}
}

@software{niji2022,
  author = {Midjourney},
  title = {Niji: Let's make magic anime pictures!},
  note = {Accessed: November 10, 2025}
}

@inproceedings{huang2024epidiff,
  title={Epidiff: Enhancing multi-view synthesis via localized epipolar-constrained diffusion},
  author={Huang, Zehuan and Wen, Hao and Dong, Junting and Wang, Yaohui and Li, Yangguang and Chen, Xinyuan and Cao, Yan-Pei and Liang, Ding and Qiao, Yu and Dai, Bo and others},
  booktitle={Proceedings of the IEEE/CVF Conference on Computer Vision and Pattern Recognition},
  pages={9784--9794},
  year={2024}
}

@inproceedings{zhang2025ar,
  title={AR-1-to-3: Single Image to Consistent 3D Object via Next-View Prediction},
  author={Zhang, Xuying and Zhou, Yupeng and Wang, Kai and Wang, Yikai and Li, Zhen and Jiao, Shaohui and Zhou, Daquan and Hou, Qibin and Cheng, Ming-Ming},
  booktitle={Proceedings of the IEEE/CVF International Conference on Computer Vision},
  pages={26273--26283},
  year={2025}
}

@article{chen2025humo,
  title={Humo: Human-centric video generation via collaborative multi-modal conditioning},
  author={Chen, Liyang and Ma, Tianxiang and Liu, Jiawei and Li, Bingchuan and Chen, Zhuowei and Liu, Lijie and He, Xu and Li, Gen and He, Qian and Wu, Zhiyong},
  journal={arXiv preprint arXiv:2509.08519},
  year={2025}
}

@article{ye2023ip,
  title={Ip-adapter: Text compatible image prompt adapter for text-to-image diffusion models},
  author={Ye, Hu and Zhang, Jun and Liu, Sibo and Han, Xiao and Yang, Wei},
  journal={arXiv preprint arXiv:2308.06721},
  year={2023}
}
}


\end{document}